\documentclass[10pt,twocolumn,letterpaper]{article}

\usepackage{cvpr}
\usepackage{times}
\usepackage{epsfig}
\usepackage{graphicx}
\usepackage{amsmath}
\usepackage{amssymb}
\usepackage{array}

\usepackage{multirow}
\usepackage{fancyhdr}
\usepackage{hyperref}
\usepackage{booktabs}
\usepackage{caption}

\usepackage{float}

\newcommand{\PreserveBackslash}[1]{\let\temp=\\#1\let\\=\temp}
\newcolumntype{C}[1]{>{\PreserveBackslash\centering}p{#1}}
\newcolumntype{R}[1]{>{\PreserveBackslash\raggedleft}p{#1}}
\newcolumntype{L}[1]{>{\PreserveBackslash\raggedright}p{#1}}

\cvprfinalcopy 

\def\cvprPaperID{****} 

\begin{document}

\title{Technical Report for SoccerNet Challenge 2022 - Replay Grounding Task}  

\author{Shimin Chen\textsuperscript{1}\thanks{These authors contributed equally to this work}
\quad
Wei Li\textsuperscript{1}$^\ast$
\quad 
Jiaming Chu\textsuperscript{1,2}$^\ast$
\quad
Chen Chen\textsuperscript{1}$^\ast$
\quad
Chen Zhang\textsuperscript{1}
\quad
Yandong Guo\textsuperscript{1} 
\\
\textsuperscript{1}OPPO Research Institute.
\quad
\textsuperscript{2}Beijing University of Posts and Telecommunications.
\\
{\tt\small \{chenshimin1, liwei19, chenchen, zhangchen4, guoyandong\}@oppo.com}
\\
{\tt\small chujiaming886@bupt.edu.cn}
}

\twocolumn[{%
\renewcommand\twocolumn[1][]{#1}%
\maketitle
\begin{center}
    \centering
    \includegraphics[height=3.85cm,width=16cm]{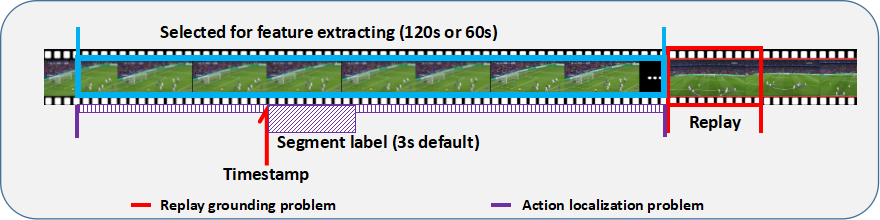}
    \captionof{figure}{Transform of problems. For every replay cut, we select a long video cut for feature extracting. And according to timestamp, we generate segment label. So the aim of replay grounding task is transformed to predict the action segment from selected long video cut. The red element in the graph is the Replay grounding task and the purple element is the Action localization task.} \label{figure:transform_of_problem}
\end{center}%

}]


\thispagestyle{empty}

\section{Method}
In order to make full use of video information, we transform the replay grounding problem into a video action location problem. We apply a unified network Faster-TAD\cite{chen2022faster} proposed by us for temporal action detection to get the results of replay grounding. Finally, by observing the data distribution of the training data, we refine the output of the model to get the final submission.

\subsection{Problem Transformation}
Replay grounding consists in retrieving the timestamp of the action shown in a given replay shot within the whole game. Action usually happens during video snips and we found that the clip after the timestamp of the action contains a lot of discriminate information, such as ``Throw-in’’. In order to make full use of video information, we transform the replay grounding problem into a temporal action detection problem. For SoccerNet replay grounding, we employ pre-extracted features as input. In the training and testing of the model, we select 120 seconds and 60 seconds video features before replay timestamps as input respectively. We set the timestamp as the starting second of the segment labels with 3 seconds length, as shown in Fig.~\ref{figure:transform_of_problem}. In this way, predicting the timestamp corresponding to replay just is predicting the starting second of Segment label. We could first apply the Faster-TAD to predict the Segment label and then regard its starting second as timestamp results in this task.

\subsection{Feature Engineering}
In order to better mine these information, we define two atomic labels. In first label definition named ``6s’’, we define atomic action segment from 2 seconds before the labeled position to 4 seconds after that. In second label definition named ``3s style 1’’, to distinguish the information before and after labeled position more clearly, we define two atomic action segments from one labeled position. One is from 3 seconds before the labeled position to labeled position, and another is labeled position to 3 seconds after that. 

\begin{figure*}[h]
\centering
\includegraphics[scale=0.6]{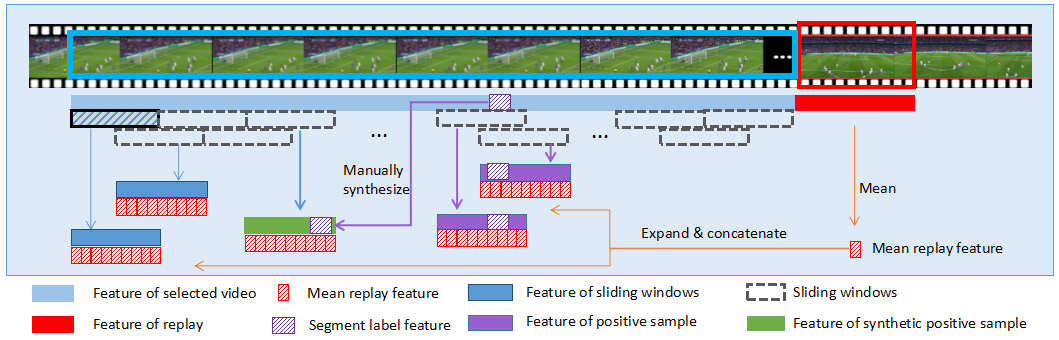}
\caption{Processing of feature. we process both features with sliding windows of $size = 16$ seconds, $stride=8$ seconds.We get the features of replay and pooling it to 1 channel as Mean Replay Features. We concatenate the Mean Replay Features with each features within sliding windows. About synthesize positive samples, we randomly choose clip features which don't include any Segment label, and cover a part of it randomly with the feature of Segment label.}
\label{figure:Feature_preprocess}
\end{figure*}

Then we train two Swin Transformer\cite{liu2021swin} as classification models to learn different action characteristics, and final extract 4fps features by windows with 16 seconds length and 8 seconds stride. And the results of the two atomic classifiers are shown in the table \ref{table:ato}.

\begin{table}[h] 
\caption{The results of the two atomic classifiers.} \label{table:ato}
\begin{tabular}{cccc}
\hline
style     & top1\_acc & top5\_acc & mean\_class\_accuracy \\ \hline
6s        & 88.69    & 98.48    & 81.89               \\
3s style1 & 73.13    & 96.19    & 70.96               \\ \hline
\end{tabular}
\end{table}

\subsection{Feature Preprocessing in TAD}

Base on the extracted features, we process both features with sliding windows of $size = 16$ seconds, $stride=8s$. We get the features of replay and pooling it to 1 channel as Mean Replay Features. We concatenate the Mean Replay Features with each features within sliding windows, as shown in Fig.~\ref{figure:Feature_preprocess}. During training, we only take the features which include the Segment label. To get more samples for training, we synthesize positive samples manually. We randomly choose clip features which don't include any Segment label, and cover a part of it randomly with the feature of Segment label. The quantity rate between synthetic positive samples and positive samples is 1:1. Then, we resize the video features to 100 along time dimension by linear interpolation. Last but not least, we adopted two-stream feature joint training, which exploits the complementary between different features. Features are combined using Auxiliary-Features Block mentioned in \cite{Chen2022FasterTADTT}. 

\subsection{Temporal Action Detection}
We apply a Faster-RCNN like network in temporal action detection, Faster-TAD. By jointing temporal proposal generation and action classification with multi-task loss and shared features, Faster-TAD simplifies the pipeline of TAD.

We construct our Faster-TAD with feature sequences extracted from raw video as inputs by Swin Transformer\cite{liu2021swin} Extractor. 

We process the feature sequences with a base module to extract shared features, which consists of a CNN Layer, a Relu Layer, and a GCNeXt\cite{xu2020g} Block. We then exert a Proposal Generation Mechanism to obtain most credible $K$ coarse proposals, where $K$ is 120. Proposals and shared features are further utilized to get more accurate boundaries by Boundary Regression Refinement Module\cite{qing2021temporal}. At the same time, shared features and proposals are employed to get the semantic labels of action instances with Context-Adaptive Proposal Module. We make some improvements to tackle the challenges in temporal action detection. 

\subsection{Implementation Details}

We train our model in a single network, with batch size of 32 on 8 gpus. The learning rate is $8\times{10}^{-4}$ for the first 3 epochs, and is reduced by 10 in epoch 3. We train the model with total 12 epochs. In inference, we apply Soft-NMS\cite{bodla2017soft} for post-processing, and select the top-M prediction for final evaluation.
\section{Results}

In the process of training and validation, our model based on positive samples with different features gets good results. With features of positive samples from ``3s style1’’, ``6s’’ and ``3s+6s’’, we get 90.84, 91.45, 92.19 in AUC, 67.69, 66.07, 70.54 in AR@1, 86.56, 86.08, 88.34 in AR@5 on validation set.

Zhao et al.\cite{zhou2021feature} is the winner of SoccerNet replay grounding 2021. 
As shown in Table \ref{tab:tab3}, with Faster-TAD network, by two-stream features which combine ``3s style1’’ and ``6s’’, we reached a tight mAP of 52.31\% in test set, bringing a gain of 26.76\% mAP. 

\begin{table}[ht]
    \centering
    \caption{Replay grounding results on test set of SoccerNet-Replay grounding, measured by tight average-mAP and loose average-mAP measured by tight average-mAP introduced by \cite{giancola2018soccernet}. }\label{tab:tab3}
    \scalebox{0.7}{
    \begin{tabular}{c|c|c|c|c|c|c}
    \hline
        Method & Feature & tight & loose & AUC & AR@1 & AR@5 \\\hline
        Zhao et al.\cite{zhou2021feature} & Baidu & 25.55 & 76.00 & - & - & - \\
        Faster-TAD(ours) & Swin(3s style1) & 48.61 & 66.64 & 90.84 & 67.69 & 86.56 \\
        Faster-TAD(ours) & Swin(6s) & 42.42 & 62.63 & 91.45 & 66.07 & 86.08  \\
        Faster-TAD(ours) & Swin(3s+6s) & 52.31 & 68.57 & 92.19 & 70.54 & 88.34 \\\hline
    \end{tabular}}
\end{table}


{\small
\bibliographystyle{unsrt}
\bibliography{egbib}
}

\end{document}